\documentclass[letterpaper, 10 pt, conference]{ieeeconf}

\IEEEoverridecommandlockouts
\usepackage{cite}

\usepackage{amsmath}
\usepackage{amsthm}
\usepackage{amssymb}

\usepackage[ruled,linesnumbered]{algorithm2e}
\SetKwComment{Comment}{/* }{ */}
\RestyleAlgo{ruled}

\theoremstyle{definition}

\newtheorem{definition}{Definition}

\theoremstyle{plain}
\newtheorem{proposition}{Proposition}

\theoremstyle{plain}
\newtheorem{lemma}{Lemma}

\theoremstyle{plain}
\newtheorem{theorem}{Theorem}

\usepackage[pagewise]{lineno}
\usepackage{graphicx}

\graphicspath{{./figures/}}
\DeclareUnicodeCharacter{2212}{\ensuremath{-}}
\usepackage{subfig}
\usepackage{color}

\title{\LARGE \bf
Spatio-temporal Motion Planning for Autonomous Vehicles with Trapezoidal Prism Corridors and B\'{e}zier Curves
}

\author{Srujan Deolasee$^{1}$, Qin Lin$^{2}$, Jialun Li$^{3}$, and John M. Dolan$^{4}$
\thanks{*This work was supported by the National Science Foundation.}
\thanks{$^{1}$Srujan Deolasee is with Department of Computer Science \& Information Systems, Birla Institute of Technology and Science, Pilani, Rajasthan, India
        {\tt\small f20191139@pilani.bits-pilani.ac.in}}%
\thanks{$^{2}$Qin Lin is with the Department of Electrical Engineering and Computer Science, Cleveland State University, Cleveland, OH, the USA
        {\tt\small q.lin80@csuohio.edu} (corresponding author)}%
        \thanks{$^{3}$Jialun Li is with Dajiang Innovation Technology Co.,  Ltd, Shenzhen, China
        {\tt\small jialunli97@gmail.com}}
\thanks{$^{4}$John M. Dolan is with the Robotics Institute, Carnegie Mellon University, Pittsburgh, PA, the USA
        {\tt\small jdolan@andrew.cmu.edu}}%
}

\begin{document}
\maketitle
\thispagestyle{empty}
\pagestyle{empty}

\begin{abstract}

Safety-guaranteed motion planning is critical for self-driving cars to generate collision-free trajectories. A layered motion planning approach with decoupled path and speed planning is widely used for this purpose. This approach is prone to be suboptimal in the presence of dynamic obstacles. Spatial-temporal approaches deal with path planning and speed planning simultaneously; however, the existing methods only support simple-shaped corridors like cuboids, which restrict the search space for optimization in complex scenarios. We propose to use trapezoidal prism-shaped corridors for optimization, which significantly enlarges the solution space compared to the existing cuboidal corridors-based method. Finally, a piecewise B\'{e}zier curve optimization is conducted in our proposed corridors. This formulation theoretically guarantees the safety of the continuous-time trajectory. We validate the efficiency and effectiveness of the proposed approach in numerical and CommonRoad simulations.

\end{abstract}

\begin{keywords}

Autonomous vehicle, motion planning, trajectory optimization

\end{keywords}

\section{INTRODUCTION}

Motion planning is one of the key modules in autonomous driving systems. The task of motion planning in a dynamic traffic environment is to generate trajectories for a low-level controller to follow considering collision-free safety constraints, dynamic feasibility, and comfort. A Frenet frame \cite{werling_optimal_2012} is commonly used for motion planning due to the significant advantage of its independence from complex road geometry. The lateral motion (in the $L$ direction) and longitudinal motion (in the $S$ direction) can be projected onto the reference, which is usually the centerline of the road with an arbitrary shape. Including the time dimension $T$, a 3D $S − L − T$ coordinate system can be established for insightful and convenient planning.

\textbf{Path-speed (or Layered planning)} is a practical real-time solution to decompose a planning problem into two stages: path planning and speed planning \cite{gu_runtime-bounded_2016, wenda_xu_real-time_2012, li_real-time_2016}. A path ($S-L$) is generated in the first stage in a static or low-speed environment. The generation of the speed profile ($S-T$ or $L-T$) in the speed planning stage allows an AV to respond to dynamic obstacles. The significant limitation of layered planning is that it is prone to be suboptimal in the presence of dynamic obstacles in complicated scenarios.

\begin{figure}[htbp]
    \centering

    \subfloat[]{\includegraphics[width=0.2\textwidth, angle=90]{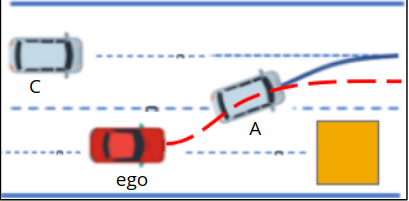}\label{fig:motivation_a}} \quad\quad
    \subfloat[]{\includegraphics[width=0.2\textwidth]{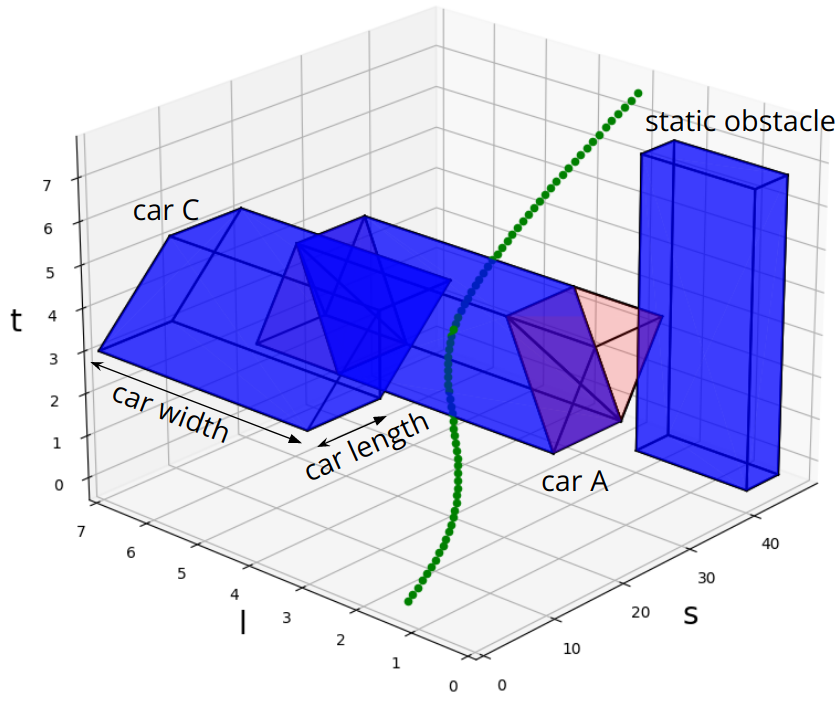}\label{fig:motivation_b}}\\
     \subfloat[]{\includegraphics[width=0.2\textwidth]{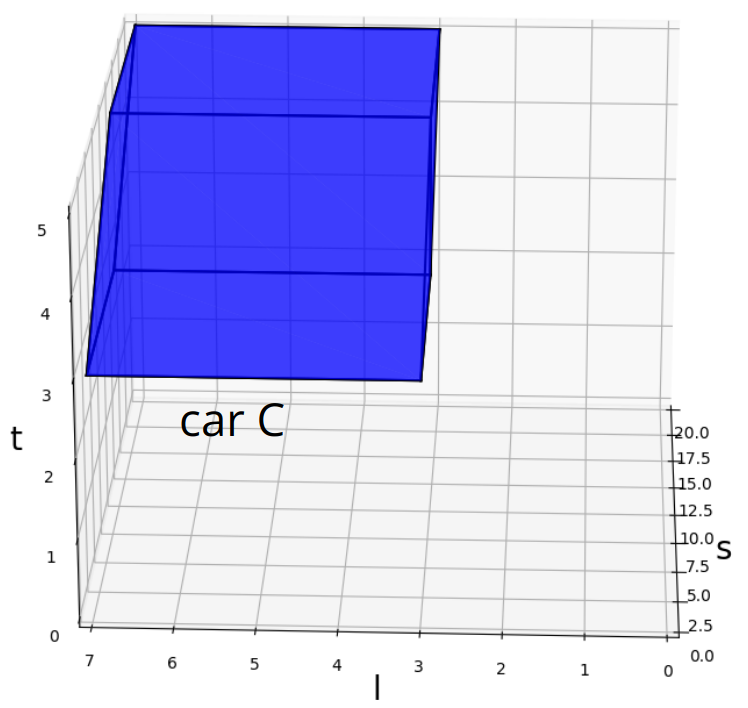}\label{fig:motivation_c}}
    \subfloat[]{\includegraphics[width=0.2\textwidth]{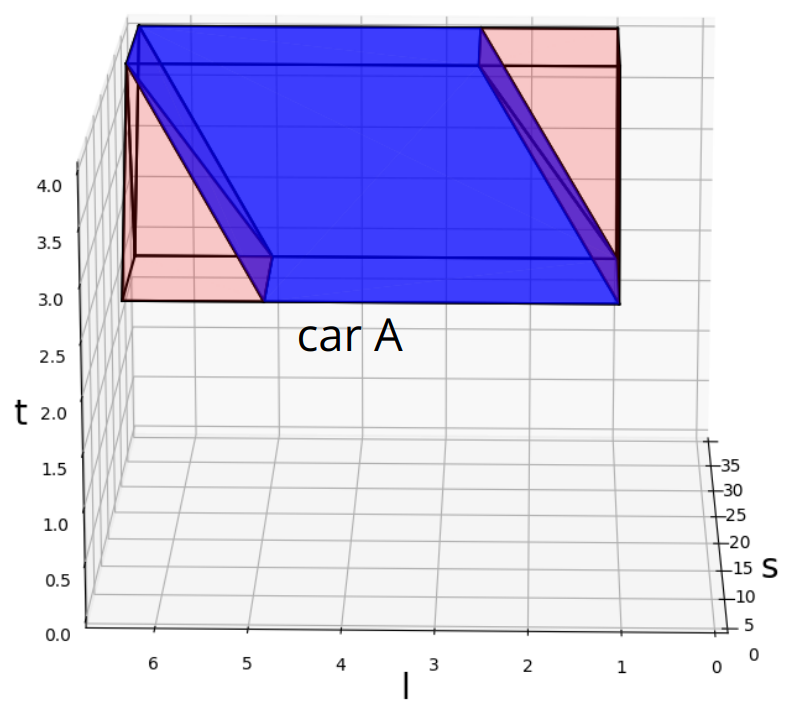}\label{fig:motivation_d}}
    \caption{Yielding example. (a) bird's-eye view; (b) our spatio-temporal planning in 3D $S-L-T$ graph; (c) another angle of view for car C; (d) another angle of view for car A.}
    \label{fig:motivation}
\end{figure}

\textbf{Spatio-temporal planning} considers spatial and temporal maneuvers simultaneously \cite{ding_safe_2019, mercy_real-time_2016, ziegler_trajectory_2014}. This method of direct optimization in the 3D $S-L-T$ space is generally superior to the layered planning approach due to the larger search space. See the motivating example illustrated in Fig. \ref{fig:motivation} (discussed in detail in Section III.A): in driving scenarios involving even small deviations along the lateral direction, coupled longitudinal and lateral planning helps guarantee global optimality. 

Ensuring safety is the vital objective of motion planning. Many existing speed planning methods use discrete time instants to impose safety constraints. However, a provable safety guarantee independent of sampling time in continuous time space is preferable. To address this problem, the spatial corridor is widely applied in trajectory generation. We are motivated by these efforts to further extend the spatial corridor to the spatio-temporal domain to cope better with dynamic obstacles. The convex hull property of B\'{e}zier polynomials is leveraged to enforce that the continuous trajectory always falls into a safe spatio-temporal region. In addition, such an optimization problem's solution space is enlarged via our proposed trapezoidal-prism-shaped corridors.

The main contributions of our work can be briefly described as follows:
\begin{enumerate}
    \item We propose an efficient convexification algorithm to construct 3D convex-feasible regions consisting of trapezoidal-prism-shaped corridors.
    \item We provide a sufficient condition on coefficients of the Bézier polynomials to theoretically guarantee the trajectory’s safety in trapezoidal-prism corridors. Compared with existing cuboidal corridors \cite{ding_safe_2019}, the condition is relaxed and the solution space is significantly enlarged, which leads to a higher chance of finding an optimal solution.
\end{enumerate}

The remainder of this paper is structured as follows. We review related works in Sec. II. We introduce necessary notations and background materials in Sec. III. The 3D convex safe region construction is presented in Sec. IV. In Sec. V, we present our optimization formulation. The simulation results and analysis can be found in Sec. VI. We make concluding remarks in Sec. VII.

\section{RELATED WORKS}

\subsection{Speed Planning}
Speed planning techniques can be classified into three categories: 1) search and optimization; 2) sampling lattices and selecting the minimum-cost trajectory; 3) approximated optimization. 		
The \textbf{Search and optimization} method searches for the best candidate speed profile and optimizes the curve for smoothness; see the post-optimization method\cite{wenda_xu_real-time_2012}, the Baidu EM motion planner \cite{fan_baidu_2018}, and the Piecewise-Jerk Speed Optimization \cite{zhou_autonomous_2021}.
The \textbf{Sampling} approach samples different speed lattices combined with path lattices. The generated local spatial-temporal trajectories are evaluated and the one with minimum cost is selected. Related works can be found in \cite{gu_runtime-bounded_2016,wenda_xu_real-time_2012,li_real-time_2016}. Most works in the first and the second categories conduct search and optimization directly in the $S-T$ graph. \textbf{Approximated optimization} considers a vehicle dynamic model in a sequential optimization problem; see convex feasible set algorithm \cite{liu_speed_2017} and optimal control methods, such as model predictive control (MPC)\cite{qian_motion_2016} and constrained iterative linear quadratic regulator (CiLQR)\cite{pan_safe_2020}. The advantage of these approaches is that they mitigate the planning and control inconsistency problem since the dynamic model has already been considered in the planning layer. However, the disadvantage is the high computation cost.

\subsection{Corridor generation for Autonomous Vehicles}
The spatial corridor is widely used in trajectory generation. Some previous works generate the corridors in a static environment and cannot deal with dynamic obstacles \cite{zhu_convex_2015}, \cite{erlien_safe_2013}.
Liu et al. \cite{liu_convex_2018} find a convex feasible set around the reference trajectory, but the computation complexity restricts the method for real-time applications. Our previous work proposes the use of trapezoidal corridors for convexifying 2D space in the $S-T$ graph \cite{li_motion_nodate}. Zhang et al. present a general convex spatio-temporal corridors-based approach \cite{zhang_sufficient_2022}. Xu et al. propose using a modified vertical cell decomposition approach for speed planning in \cite{xu_speed_nodate}. All these methods suffer from the limitations of the layered planning approach discussed in the previous section. Ding et al. use the spatio-temporal semantic corridor (SSC) method to uniformly express obstacles and traffic rules in the 3D $S-L-T$ space \cite{ding_safe_2019}. However, restricting the shape of the corridors to simple cuboids drastically limits the search space for optimization in complex scenarios. Our proposed method of extending trapezoid-shaped 2D corridors in $S-T$ to 3D $S-L-T$ space significantly enlarges the solution space for trajectory optimization. This enables us to extend the spatial corridor to the spatio-temporal domain to cope with dynamic obstacles while meeting the real-time requirement.

\subsection{Bézier Polynomials-Based Planning}

Previously, monomial basis polynomials have been used to generate trajectories \cite{werling_optimal_2012}, \cite{fan_baidu_2018}. However, these methods often fail to represent highly constrained maneuvers in the presence of dynamic obstacles. They also fail to give safety guarantees between sample points, as the constraints are only enforced/checked on a finite set of sampled points. In \cite{gonzalez_speed_2016}, a smooth and continuous speed profile is computed by proper curve concatenation without optimization and dynamic obstacles. B\'{e}zier polynomials combined with rectangular corridors was originated in the area of unmanned aerial vehicles (UAVs) \cite{gao_online_2018}, \cite{zhou_robust_2019}. Ding et al. extended it for motion planning of unmanned ground vehicles (UGVs) \cite{ding_safe_2019}. The significant limitation is that the proposed cuboidal corridor representation fails to make the most of free space for optimization. In our work, we propose to use time-dependent trapezoidal prism-shaped corridors and give sufficient conditions to enforce B\'{e}zier curves in these time-dependent corridors for safety. It is theoretically proved that the trapezoidal prism-shaped corridors can enlarge the solution space for improved optimization.

\section{$S-L-T$ GRAPH AND TRAJECTORY REPRESENTATION}

In this section, we briefly introduce background materials on the $S-L-T$ graph, B\'{e}zier polynomials, and trajectory representations using piecewise B\'{e}zier polynomials.

\subsection{Representing Dynamic Agents in $S-L-T$ graph}
The $S − L − T$ graph represents all traffic participants’ positions at each time step including the past, current, and prediction.
 As an example, in Fig. \ref{fig:motivation_a}, we take the case of two cars moving at constant speeds for simplicity. The scenario is described as follows: car A and the ego vehicle are driving in a lane which has a static obstacle (e.g., a construction site). A lane change maneuver is enforced for both vehicles. We list the following typical entities:
 
 1) \textbf{Static obstacle (6-zero-slope-faces type):} As the most simple entity, all the six faces have zero slopes, see the bird's-eye view of the yellow block in Fig. \ref{fig:motivation_a} and the cuboid in the $S-L-T$ plot in Fig. \ref{fig:motivation_b}.
 
 2) \textbf{Moving car with only longitudinal motion (4-zero-slope-faces type):} Car C moving straight forward is an example shown in Fig. \ref{fig:motivation_a}. The top and bottom faces of Car C in Fig. \ref{fig:motivation_b} have zero slopes. The two faces (see another angle of view for left and right faces in Fig. \ref{fig:motivation_c}) are perpendicular to the $S-L$ plane without slopes. The side length of the parallelogram along the $L$ axis is the width of the vehicle plus the safety region. The side length of the parallelogram along the $S$ axis is the length of the vehicle plus half the length of the ego vehicle as a safety region.
 
 3) \textbf{Moving car with longitudinal and lateral motions (2-zero-slope-faces type):} Car A moving left and forward is an example shown in Fig. \ref{fig:motivation_a}. As we can see in Fig.\ref{fig:motivation_b}, only the top and down faces are zero-slope.

 The 3D free space in the $S-L-T$ graph is non-convex in general. We propose an over-approximation of the \emph{2-zero-slope-faces type} parallelepiped, e.g., car A in Fig. \ref{fig:motivation_b}, into a \emph{4-zero-slope-faces type} parallelepiped, see the pink inflated space in Fig. \ref{fig:motivation_b} and Fig. \ref{fig:motivation_d}. There are two significant benefits of doing so: 1) we will show that in the presence of such parallelepipeds, we can extend our 2D corridor construction algorithm \cite{li_motion_nodate} to construct 3D trapezoidal prism-shaped corridors efficiently; 2) though we pay the cost of losing some space due to over-approximation, the safety corridor is still significantly larger than the state-of-the-art cuboidal corridors. Note that the transformation between 2D and 3D in our method is without the loss of search space. Thus, in summary, we make a good trade-off between exactness and efficiency. The faces chosen for the over-approximation step are decided by a simple minimization of the volume of search space compromised in the process. Intuitively, if the lateral velocity is less than the longitudinal velocity of the vehicle, the corresponding faces are chosen for over-approximation.

\subsection{B\'{e}zier Polynomials and Properties}

A B\'{e}zier polynomial is a polynomial function represented by linear combinations of Bernstein bases. The $n$th-order B\'{e}zier polynomial is written as
$$
B(t)=c_{0} b_{n}^{0}(t)+c_{1} b_{n}^{1}(t)+\cdots+c_{n} b_{n}^{n}(t)=\sum_{i=0}^{n} c_{i} b_{n}^{i}(t)
$$
where the Bernstein bases satisfy $b_{n}^{i}(t)=C_{i}^{n} \cdot t^{i} \cdot(1-$ $t)^{n-i}, t \in[0,1]$. The coefficients of the polynomial $c_{i}(i=$ $0,1, \ldots, n$) are also called control points. Compared to monomial polynomials, B\'{e}zier curves have the following properties:

\begin{itemize}
    \item The time interval is defined on $t \in[0,1]$.
    \item The Bézier polynomial starts at control point $B(0)=c_{0}$ and ends at $B(1)=c_{n}$.
    \item Convex hull property: The Bézier curve $B(t)$ is confined within the convex hull of control points.
    \item Hodograph property: By the hodograph property, the derivative of $B(t), \dot{B}(t)$, can also be written as a Bézier polynomial with control points $c_{i}^{1}=n \cdot\left(c_{i+1}-c_{i}\right), i=$ $0,1, \ldots, n-1$. By applying the convex hull property to the derivative Bézier curve, the entire dynamical profile of the original curve $B(t)$ can be confined within a given dynamical range.
\end{itemize}

\subsection{Trajectory Representation using B\'{e}zier Polynomials}
To mitigate the numerical instability issue, piecewise B\'{e}zier polynomials with lower orders are used instead of using a high-order B\'{e}zier polynomial for the whole planning horizon. Each piece of the trajectory is associated with one trapezoidal-prism corridor. Note that $B(t)$ is defined on a fixed time interval $[0,1]$. For a whole trajectory with $m+1$ pieces, in each piece $\left[T_{k}, T_{k+1}\right] (k = 0,1,\ldots,m)$, we use a scaling transformation and translation transformation in the time domain to map it into the interval $[0,1]$ [13]. Then, the whole piece-wise trajectory in one dimension $\sigma\in\{s,l\}$ is:
$$
f^{\sigma}(t)=\left\{\begin{array}{c}
h_{0} B_{0}\left(\frac{t-T_{0}}{h_{0}}\right), t \in\left[0, T_{1}\right] \\
h_{1} B_{1}\left(\frac{t-T_{1}}{h_{1}}\right), t \in\left[T_{1}, T_{2}\right] \\
\vdots \\
h_{m} B_{m}\left(\frac{t-T_{m}}{h_{m}}\right), t \in\left[T_{m}, T_{m+1}\right] .
\end{array}\right.
$$
where $h_i$ is the scaling transformation factor and $T_i$ is the translation transformation factor for $i = 0,1,\ldots,m$ with setting $T_0 = 0$.

\section{CORRIDOR GENERATION}

In this section, a convexification algorithm is introduced to construct convex corridors from the original non-convex optimization problem for real-time solving. A reference trajectory is often used to provide a warm start to the optimization process. In this work, we use simple piecewise functions for generating valid reference waypoints in the configuration space of the ego vehicle.

\subsection{Piecewise Convex Safe Regions Representations}

Suppose the whole safe region is divided into $m+1$ pieces with time intervals $\left[T_{0}, T_{1}\right], \ldots,\left[T_{m}, T\right]$ and $T =T_{m+1}$, with each interval corresponding to a convex safe region. The details of such a convexification algorithm will be introduced in the next section. The $k$-th convex safe region in $S-L-T$ space can be represented as 
$$\mathcal{S}_{k} = \{\ (t_i, s_i, l_i)\ | $$
$$\underline{p^{k}_0} + h_k \underline{p^{k}_1} \frac{t_i - T_k}{h_k} \leq s_i \leq \overline{p^{k}_0} + h_k \overline{p^{k}_1} \frac{t_i - T_k}{h_k},$$
$$l_{beg} \leq l_i \leq l_{end},\ t_i \in [T_k, T_{k+1}]\ \}$$
where $s_i$ and $l_i$ are the longitudinal and lateral coordinates of the $i^{th}$ control point respectively, $\underline{p_{0}^{k}}, \underline{p_{1}^{k}}$ are bias and skew of the lower bound and $\overline{p_{0}^{k}}, \overline{p_{1}^{k}}$ are those of the upper bound. $h_{k}$ denotes the length of the $k$-th time interval and satisfies $h_{k}=T_{k+1}-T_{k}, k=0,1, \ldots, m$.

Then, the whole safe region is the union of a set of piecewise-safe sub-regions: $
\mathcal{S}=\mathcal{S}_{0} \cup \cdots \cup \mathcal{S}_{m}$. The speed planning is safe if $\forall t_{0} \in[0, T], s\left(t_{0}\right) \in \mathcal{S}, l\left(t_{0}\right) \in \mathcal{S}$, which is equivalent to for $t_{0} \in\left[T_{k}, T_{k+1}\right], s\left(t_{0}\right) \in \mathcal{S}_{k}, l\left(t_{0}\right) \in \mathcal{S}_{k}, k=0,1, \ldots, m$, i.e.,
$$
\underline{p_{0}^{k}}+h_{k} \underline{p_{1}^{k}} \frac{t_{0}-T_{k}}{h_{k}} \leq s\left(t_{0}\right) \leq \overline{p_{0}^{k}}+h_{k} \overline{p_{1}^{k}} \frac{t_{0}-T_{k}}{h_{k}}
$$
$$
l_{beg} \leq l\left(t_{0}\right) \leq l_{end}
$$

\subsection{Construction of Piecewise-Convex Safe Regions}
\begin{figure}[htbp]
    \centering
    \subfloat[$S-L-T$ graph]{\includegraphics[width=0.2\textwidth]{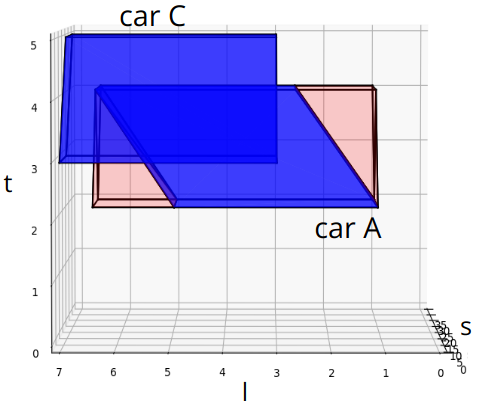}\label{fig:f5}} \quad\quad
    \subfloat[over-approximation (red)]{\includegraphics[width=0.22\textwidth]{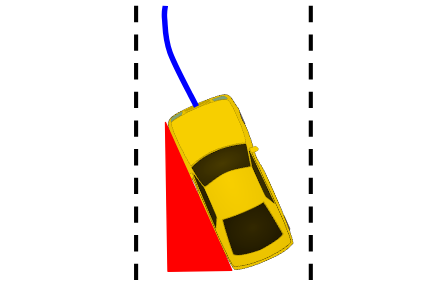}\label{fig:f5.5}}\\
    \subfloat[$S-T$ graph for $l \in [1,3)$]{\includegraphics[width=0.2\textwidth]{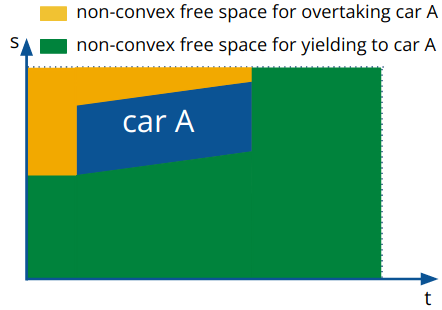}\label{fig:f7}}\quad
    \subfloat[$S-T$ graph for $l \in [3,6.7)$]{\includegraphics[width=0.2\textwidth]{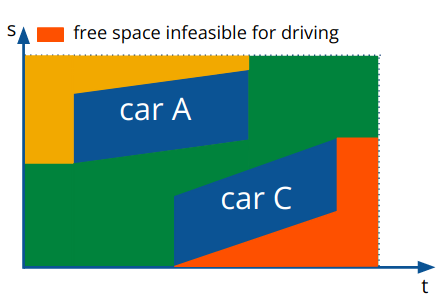}\label{fig:f6}}
    \caption{Scenario in Fig. \ref{fig:motivation_b} after over-approximating car A}
    \label{fig:Fig 2}
\end{figure}
Algo. 1 outlines the 3D trapezoidal corridor generation process. The original non-convex space is sliced along the $L$ axis at the starting or ending $L$ coordinates of any obstacles in the $S-L-T$ graph. This generates 3D chunks of the non-convex space which can be projected in a 2D $S-T$ graph \textit{without} the loss of any search space. As an example, in Fig. \ref{fig:f5}, any slices at $L$ coordinates in the range $[1,3)$ will give us the 2D $S-T$ cross-section as seen in Fig. \ref{fig:f7}. Similarly, any slice at an $L$ coordinate between $[3, 6.7)$ will give us the $S-T$ graph as seen in Fig. \ref{fig:f6}. The inputs to Algo. 1 are 
the upper and lower bounds in the $S$ and $L$ direction w.r.t. the ego vehicle and the road. The bounds are measured over a time horizon using a discrete time interval $\Delta$.
For each slice, we construct 2D convex corridors in the corresponding $S-T$ graph. In this work, we extend the 2D convex trapezoidal corridor generation algorithm in our previous work \cite{li_motion_nodate}. The new algorithm is presented as Algo. 2.

Algo. 2 outlines the construction of 2D piecewise-convex safe regions in any given $S-T$ cross-section along the $L$ axis. The lower and upper bounds in the $S$ direction serve as inputs to this algorithm. We refer readers to \cite{li_motion_nodate} for further details about the working of Algo. 2.
A key modification in our work is that in the subroutine $SingleRegionCaculate()$, we also initialize the upper and lower boundaries of the regions in the $L$ direction (Algo. 3. Lines 6,7). Our over-approximation step and the design of Algo. 1 guarantee that these boundaries are the same for all 2D convex regions generated by Algo. 2. Thus, we essentially get 2D trapezoidal-shaped corridors \textit{dragged} along the $L$ axis to form 3D trapezoidal prism-shaped convex corridors. Finally, $RegionSplit()$ is used to check the length of each 2D convex region. If it is above a user-defined threshold (e.g., $1\ s$ in our experimental setting), it will be split into multiple sub-regions, for which the time intervals are all below the threshold. This refinement operation aims to avoid underfitting.

This 2D corridor generation process is repeated for all distinct obstacle boundaries in our $S-L-T$ graph (Line 1, Algo. 1). The initialization of bounds along the $L$ axis ensures that we get 3D trapezoidal-shaped convex corridors. Since the length of the corridors in the $L$ direction is given by the starting or ending of the obstacles in the $S-L-T$ space, we can guarantee the safety of all the corridors generated using Algo. 1. Note that for the space divided by obstacles, we select the unique space enclosing the 
reference trajectory using the $SelectCorridors()$ method. For yielding to car A, the corridors lying in the green region of Fig. \ref{fig:f7} and \ref{fig:f6} are chosen according to the 3D reference waypoints.
Note that once the corridors are selected, the whole optimization is solved as a single problem and not decomposed into individual optimizations for separate corridors.




\begin{algorithm}[htbp]
\caption{Piecewise 3D Convex Regions Generation}
\setcounter{AlgoLine}{0}
\LinesNumbered
\KwIn{$obs, lb_s[obs], up_s[obs], lb_l[obs], up_l[obs], nums, \Delta$}

\KwOut{3D\_Region}

$\textbf{Initialize:}$ \For{$i \gets 0 \textbf{ to } obs$}{
    $corridor = Convexify2D(lb_s[i], ub_s[i], lb_l[i], ub_l[i], nums, \Delta)$;
    
    $new\_corridor.append(corridor)$;
}
$final\_corridor = SelectCorridors(new\_corridor)$;

$\textbf{Return } final\_corridor$
\end{algorithm}

\begin{algorithm}[htbp]
\small
\caption{Convexify2D}\label{alg:two}
\setcounter{AlgoLine}{0}
\LinesNumbered
\KwIn{ $lb_s, ub_s, lb_l, ub_l, nums, \Delta$}

\KwOut{ $regions$}

$\textbf{Initialize: } regions[0], i = 0, j = 1$ \Comment{i and j are counters for meta-pieces and resulting convex regions, respectively}
$SingleRegionCalculate(region,0,lb_s[0],ub_s[0],lb_s[1],$\\\ \ $ub_s[1], lb_l, ub_l)$\\
$regions.append(region)$ 
\\
\For{$i \gets 2 \textbf{ to } nums-1$}{
$lskew = (lb_s[i] - lb_s[i-1])/\Delta$ \Comment{lower bound's skew for two consecutive meta-pieces}
$uskew = (ub_s[i]-ub_s[i-1])/\Delta$ \Comment{upper bound's skew}
     \If{$||lskew-regions[j-1].lskew|| > \epsilon \textbf{ or } ||uskew-regions[j-1].uskew|| > \epsilon$}{
    $regions[j-1].t_{end} = i$\\
    $regions[j-1].t = (regions[j-1].t_{end}-regions[j-1].t_{beg})*\Delta$\\
    $SingleRegionCalculate(region,j,lb_s[i],ub_s[i],$\\\ \ $lb_s[i+1],ub_s[i+1], lb_l, ub_l)$\\
    $regions.append(region)$\\
    $j \gets j+1$
  }
}
$regions[j-1].t_{end} = nums-1$\\
$regions[j-1].t = (regions[j-1].t_{end} - regions[j-1].t_{beg})*\Delta$\\
$RegionSplit(regions)$\\
\textbf{Return} {$regions$}
\end{algorithm}
\begin{algorithm}[htbp]
\caption{Single Region Caculate}
\setcounter{AlgoLine}{0}
\LinesNumbered
\KwIn{ $region,j,lb_s[i],ub_s[i],lb_s[i+1],ub_s[i+1], lb_l, ub_l$}
\KwOut{ $region$} 
 $region.t_{beg} = j$\\
 $region.lskew = (lb_s[i + 1] - lb_s[i])/\Delta$\\
 $region.lbias = lb_s[i]$\\
 $region.uskew = (ub_s[i + 1] - ub_s[i])/\Delta$\\
 $region.ubias = ub_s[i]$\\
 $region.l_{beg} = lb_l$\\
 $region.l_{end} = ub_l$\\
\end{algorithm}

\section{PIECEWISE BÉZIER POLYNOMIAL OPTIMIZATION}
In this section, we discuss more about the limitations of using the cuboidal corridors. We then discuss the safety enforcement in our trapezoidal prism-shaped corridors. The formulation of quadratic optimization using the newly designed convex solution space is introduced thereafter.

\subsection{Limitations of Safety Enforcement in Cuboidal Corridors}
As discussed in Sec. III C, the convex hull property of the B\'{e}zier curves is used to enforce that the trajectory in the $S-L-T$ graph stays in the safe region $\mathcal{S}$. We first formally define a corridor for our trajectory generation:

\begin{definition}
\textit{Let the coefficients of the Bézier Polynomial be $c_i \in \Omega, i=0,1,\ldots,n$. Each control point has two dimensions - $\{S, L\}$. These control points lying in the safe region $\mathcal{S}$ form a subset $\mathcal{S}^{cor} \subseteq \mathcal{S}$, which is called a corridor.}
\end{definition}

Ding et al. presented the construction of cuboidal corridors in the $S-L-T$ graph \cite{ding_safe_2019}. Constraints of the control points of cuboidal corridors are given by the following proposition:

\begin{proposition}
If a trajectory has control points in each time interval satisfying $c^{k}_i \in \Omega^{k}_{cub} = \{c^k | \underline{p_{0}^{k}}+h_{k} \underline{p_{1}^{k}} \leq c^{k,s} \leq \overline{p_{0}^{k}}, l^{k}_{beg} \leq c^{k,l} \leq l^{k}_{end}, i=0,1,\ldots,n, k=0,1,\ldots,m\}, f(t)$ is guaranteed to be safe, and the upper bounds and lower bounds form cuboidal corridors $\mathcal{S}^{cub}$.
\end{proposition}
The proof of safety enforcement in rectangular corridors can be found in \cite{li_motion_nodate}, and can be straightforwardly extended to the third dimension $L$ for cuboidal corridors. 


The optimization fails if any lower bound ($\underline{p}_{0}^{k}+h_{k} \underline{p_{1}^{k}}$) is greater than the upper bound ($\overline{p_{0}^{k}}$). In order to avoid this, the time interval of the $k$-th corridor must satisfy $h_{k} \leq \frac{\overline{p_{0}^{k}}-p_{0}^{k}}{\underline{p_{1}^{k}}}$.

In \cite{ding_safe_2019}, Ding et al. propose a seed generation and cube inflation method to adjust time intervals. However, this method generates a significant number of corridors and optimized variables, which leads to a high computation cost. 
In common driving scenarios (Fig. \ref{fig:motivation}), we have $\underline{p_{1}^{k}} > 0$ or $\overline{p_{1}^{k}} > 0$, due to which the cuboidal corridors fail to cover all the safe regions. As a result, the search space is sub-optimal and the constraints on control points to enforce the trajectory 
are overtightened.

\subsection{Safety Enforcement in Trapezoidal-Prism Corridors}
The sufficient conditions of control points $c_i$ to keep the longitudinal and the lateral trajectory safe and in our proposed trapezoidal-prism corridors are built upon the following lemma.

\begin{lemma}
Let $M \in \mathbb{R}^{(n+1) \times(n+1)}$ denote the transition matrix from the Bernstein basis $\left\{b_{n}^{0}(t), b_{n}^{1}(t), \ldots, b_{n}^{n}(t)\right\}$ to the monomial basis $\left\{1, t, t^{2}, \ldots, t^{n}\right\}$. We have $M_{i, 0}=1,0 \leq$ $M_{i, j} \leq 1, i=0,1, \ldots, n, j=0,1, \ldots, n$.
\end{lemma}
The proof can be found in \cite{li_motion_nodate}. We leverage the following theorem meant for 2D trapezoidal corridors to construct 3D trapezoidal prism-shaped corridors.
\begin{theorem}
For a trajectory, if it has control points in each time interval satisfying $c_{i}^k \in \Omega^k$, where $\Omega^k = \{c^k | \underline{p_{0}^{k}}+h_{k}\underline{p_{1}^{k}}M_{i,1} \underline{p_{1}^{k}} \leq c_{i}^{k,s} \leq \overline{p_{0}^{k}}+h_{k}\overline{p_{1}^{k}}M_{i,1}, l^{k}_{beg} \leq c_{i}^{k,l} \leq l^{k}_{end}, i=0,1,\ldots,n,k=0,1,\ldots,m\}$, $f(t)$ is guaranteed to be safe. The upper and lower bounds in the $S$ and $L$ directions help form a trapezoidal prism-shaped corridor $\mathcal{S}^{trp}$.
\end{theorem}
The proof for the 2D case of the above theorem can be found in \cite{li_motion_nodate}, and can be easily extended to another dimension $L$. 

In Theorem 1, conditions on $c_{i}^{s}$ are $\underline{p_{0}^{k}}+h_{k}\underline{p_{1}^{k}}M_{i,1} \underline{p_{1}^{k}} \leq c_{i}^{k,s} \leq \overline{p_{0}^{k}}+h_{k}\overline{p_{1}^{k}}M_{i,1}$. Compared to the safety enforcement in cuboidal corridors in Proposition 1, we have $\underline{p_{0}^{k}}+h_{k}\underline{p_{1}^{k}}M_{i,1} \leq \underline{p_{0}^{k}}+h_{k} \underline{p_{1}^{k}}$ and $\overline{p_{0}^{k}}+h_{k}\overline{p_{1}^{k}}M_{i,1} \geq \overline{p_{0}^{k}}$. The advantage of having trapezoidal corridors is twofold: i) By the proof of $\underline{p_{0}^{k}}+h_{k}\underline{p_{1}^{k}}M_{i,1} \underline{p_{1}^{k}} \leq c_{i}^{k,s} \leq \overline{p_{0}^{k}}+h_{k}\overline{p_{1}^{k}}M_{i,1}$, the lower boundaries are guaranteed to be smaller than the upper boundaries all the time. Recall that for the rectangular corridors, we need to always check $h_{k} \leq \frac{\overline{p_{0}^{k}}-p_{0}^{k}}{\underline{p_{1}^{k}}}$; ii) The constraints are relaxed, therefore the solution space is enlarged compared with the rectangular corridors (see the illustration for the comparison in Fig. 3).

\subsection{Trajectory Optimization Formulation}
The objective function is established as 
\begin{equation}
 \begin{aligned}
&J = J_s + J_l\\
&J_s = w_{1} \int_{0}^{T}\left(s(t)-s^{r}(t)\right)^{2} \mathrm{~d} t+w_{2} \int_{0}^{T}\left(\dot{s}(t)-v_{s}^{r}\right)^{2} \mathrm{~d} t \\
&+w_{3} \int_{0}^{T} \ddot{s}(t)^{2} \mathrm{~d} t+w_{4} \int_{0}^{T} \dddot{s}(t)^{2} \mathrm{~d} t+w_{5}\left(s(T)-s^{r}(T)\right)^{2}\\
&J_l = w_{6} \int_{0}^{T}\left(l(t)-l^{r}(t)\right)^{2} \mathrm{~d} t+w_{7} \int_{0}^{T}\left(\dot{l}(t)-v_{l}^{r}\right)^{2} \mathrm{~d} t \\
&+w_{8} \int_{0}^{T} \ddot{l}(t)^{2} \mathrm{~d} t+w_{9} \int_{0}^{T} \dddot{l}(t)^{2} \mathrm{~d} t+w_{10}\left(l(T)-l^{r}(T)\right)^{2}\\
\end{aligned}  
\end{equation}
where $s^{r}(t)$ and $l^{r}(t)$ are the reference longitudinal and lateral trajectories, and $v^{r}_s$ and $v^{r}_l$ are the reference velocities in the two directions. For $J_s$ and $J_l$, their first terms penalize the deviation from the reference; the second ones penalize the deviation between the actual and reference speed; the third and the fourth terms penalize acceleration and jerk, respectively. The last terms penalize the deviation of the ending position from the reference. We used Optuna\cite{optuna_2019} for tuning all the 10 parameters. 

The optimization considers the following constraints:
\begin{itemize}
    \item Boundary Constraints: The piecewise curve starts from fixed position, speed, and acceleration, i.e., 
    $$
    c_{i}^{0, l} h_{k}^{(1-l)}=\left.\frac{\mathrm{d}^{l} f(t)}{\mathrm{d} t^{l}}\right|_{t=0}, \quad l=0,1,2,
    $$
    where $c_{i}^{k, l}$ is the control point for the $l$th-order derivative of the $k$-th Bézier curve. Note that $c_{i}^{k, l}$ has two dimensions: $\{S,L\}$.
    \item Continuity Constraints: The piecewise curve must be continuous at the connected time points for position, speed, and acceleration.
    $$
    c_{n}^{k, l} h_{k}^{(1-l)}=c_{0}^{k+1, l} h_{k+1}^{(1-l)}, l=0,1,2, k=0,1, \ldots, m-1
    $$
    \item Safety Constraints: With our proposed trapezoidal-prism corridors, safety constraints for the longitudinal dimension of the control point can be given as
    $$
    \underline{p_{0}^{k}}+h_{k} \underline{p_{1}^{k}} M_{i, 1} \leq c_{i}^{k, 0} \leq \overline{p_{0}^{k}}+h_{k} \overline{p_{1}^{k}} M_{i, 1}, k=0,1, \ldots, m
    $$
    and those for the lateral dimension of the control point can be given as
    $$
    l_{beg} \leq c_{i}^{k, 0} \leq l_{end}
    $$
    \item Physical Constraints: The physical constraints under consideration include the limit of a vehicle’s velocity, acceleration, and jerk. We can use the hodograph property of a Bézier curve to calculate velocity, acceleration, and jerk. The constraints are given by
    $$
    \begin{aligned}
    \beta^{k, 1} & \leq c_{i}^{k, l} \leq \overline{\beta^{k, 1}} \\
    \beta^{l} & \leq c_{i}^{k, l} \leq \overline{\beta^{l}}, l=2,3
    \end{aligned}
    $$
    where $k=0,1, \ldots, m$ and it follows that $c_{i}^{k, l+1}=(n-$ $l)\left(c_{i+1}^{k, l}-c_{i}^{k, l}\right)$. The upper bounds $\overline{\beta^{k, 1}}$ are determined by speed limits on road and centripetal acceleration constraints. Let $a_{c m}$ be the maximum acceleration permitted and $\kappa_{k}$ the maximum curvature of the path for $t \in\left[T_{k}, T_{k+1}\right]$ (see \cite{8916917} for details). The lateral acceleration constraints are given by
    $$
    c_{i}^{k, l} \leq \overline{\beta^{k, 1}}=\sqrt{\frac{a_{c m}}{\kappa_{k}}} .
    $$
The bounds on longitudinal and lateral accelerations and jerks are constant for different pieces of speed profiles.
\end{itemize}
Then, the trajectory optimization process can be formulated as a quadratic programming (QP) problem as
    $$
    \begin{aligned}
    \mathbf{P}:  \quad \min _{\mathbf{c}} \ & \frac{1}{2} \mathbf{c}^{T} \mathbf{Q}_{\mathbf{c}} \mathbf{c}+\mathbf{q}_{\mathbf{c}}^{T} \mathbf{c}+\mathrm{const} \\
    \text { s.t. } & \mathbf{A}_{e q} \mathbf{c}=\mathbf{b}_{e q} \\
    & \mathbf{A}_{i e} \mathbf{c} \leq \mathbf{b}_{i e} .
    \end{aligned}
    $$
We refer readers to the appendix of our previous work \cite{extended} for the detailed formulation process. This problem can be solved in real-time by a modern solver such as OSQP\cite{osqp}.

\section{SIMULATIONS AND RESULTS ANALYSIS}
Our framework has been implemented using C++11. All simulations are carried out on a personal computer with a 2.60GHz Intel i10-10750H processor.
\subsection{Numerical Simulations}
We conduct numerical simulations to compare the proposed approach with cuboidal corridors \cite{ding_safe_2019}. The planning horizon is 7 s. Different road scenarios are as follows:

\subsubsection{Merging into another lane due to road construction} Consider the scenario in Fig. \ref{fig:motivation}. We project different stations of the vehicles onto the $S-L-T$ graph. The initial velocity and the acceleration of the ego vehicle are $v_{s}(0) = 7.0\ \textnormal{m/s}, v_{l}(0) = 0\ \textnormal{m/s}$ and $a_{s}(0) = 0\ \textnormal{m/s\textsuperscript{2}}, a_{l}(0) = 0\ \textnormal{m/s}\textsuperscript{2}$, respectively. 

\begin{figure}[htbp]
    \centering
    \subfloat[]{\includegraphics[width=0.2\textwidth]{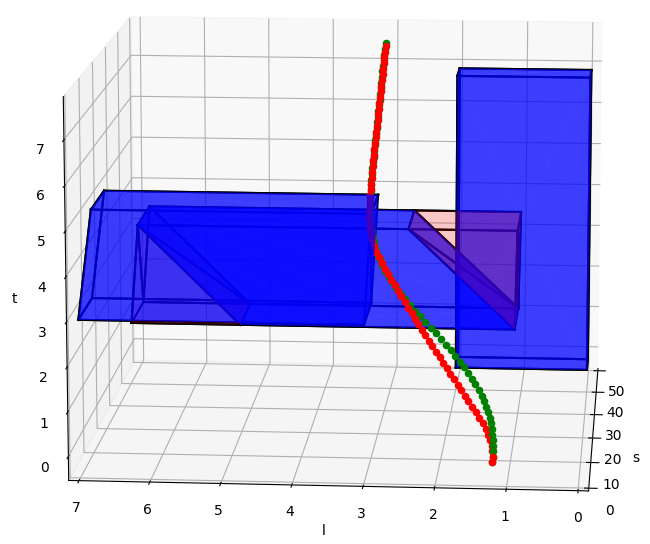}\label{fig:f8}}
    \subfloat[]{\includegraphics[width=0.2\textwidth]{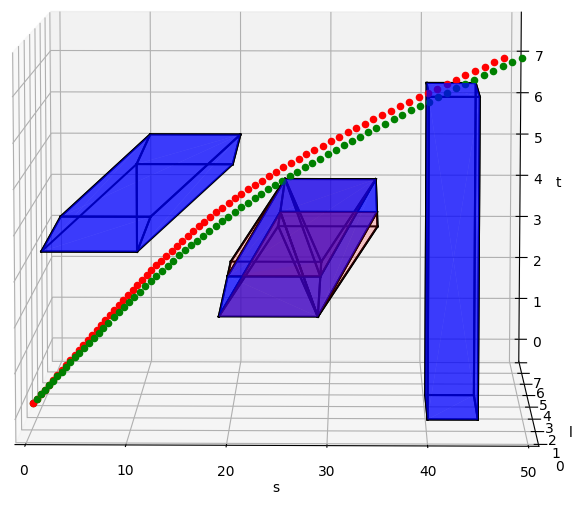}\label{fig:f9}}\\
    \textnormal{Generated trajectory using both methods (green: trapezoidal, red: cuboidal)}
    \subfloat[Longitudinal Acceleration profiles]{\includegraphics[width=0.2\textwidth]{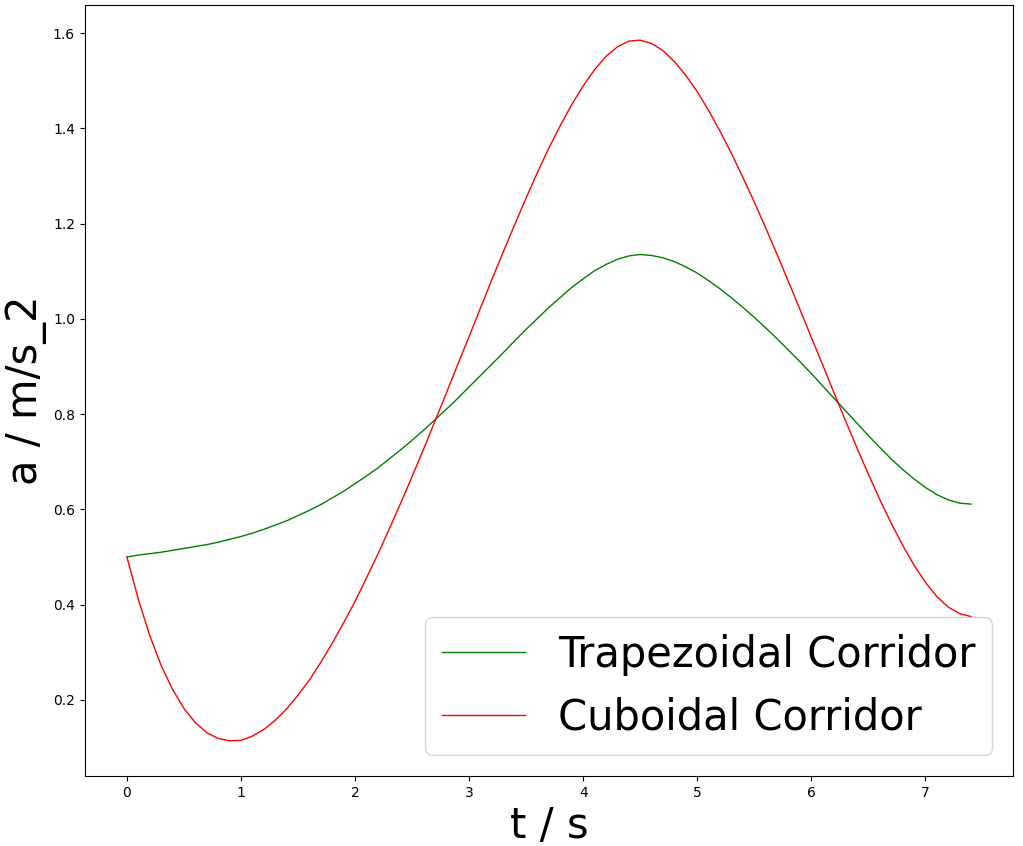}\label{fig:f12}}
    \subfloat[Lateral Acceleration profiles]{\includegraphics[width=0.2\textwidth]{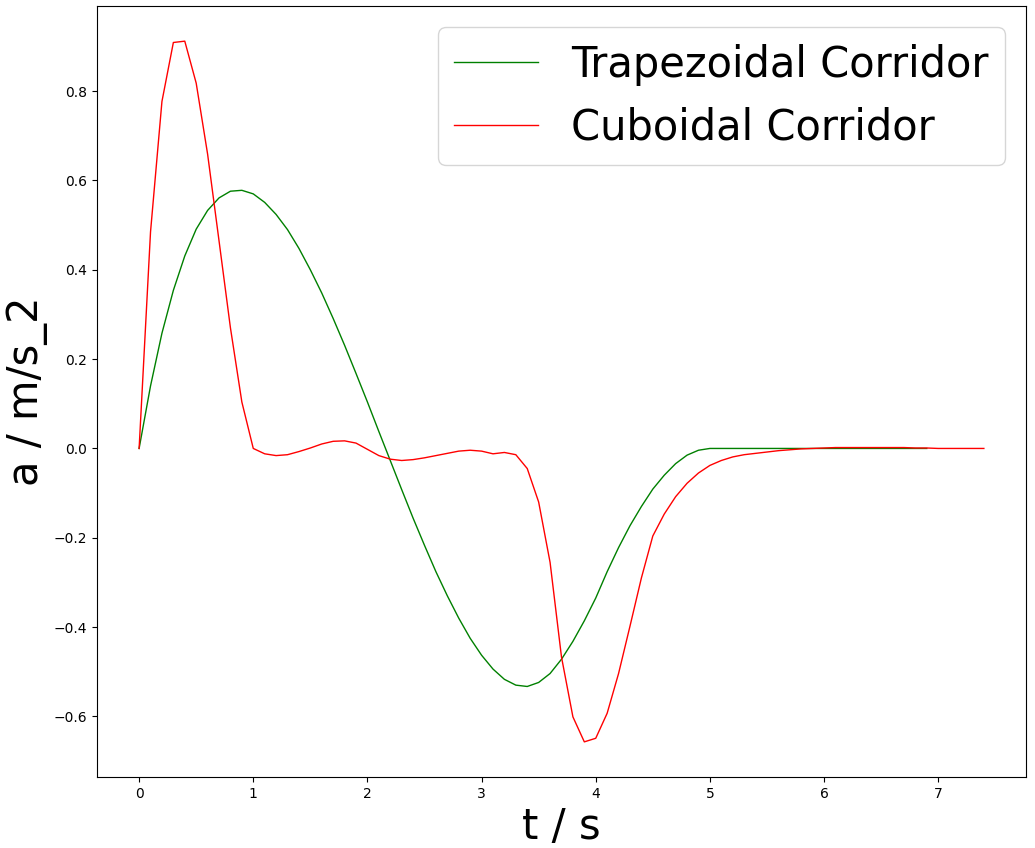}\label{fig:f13}}\\
    \subfloat[Longitudinal Velocity profiles]{\includegraphics[width=0.2\textwidth]{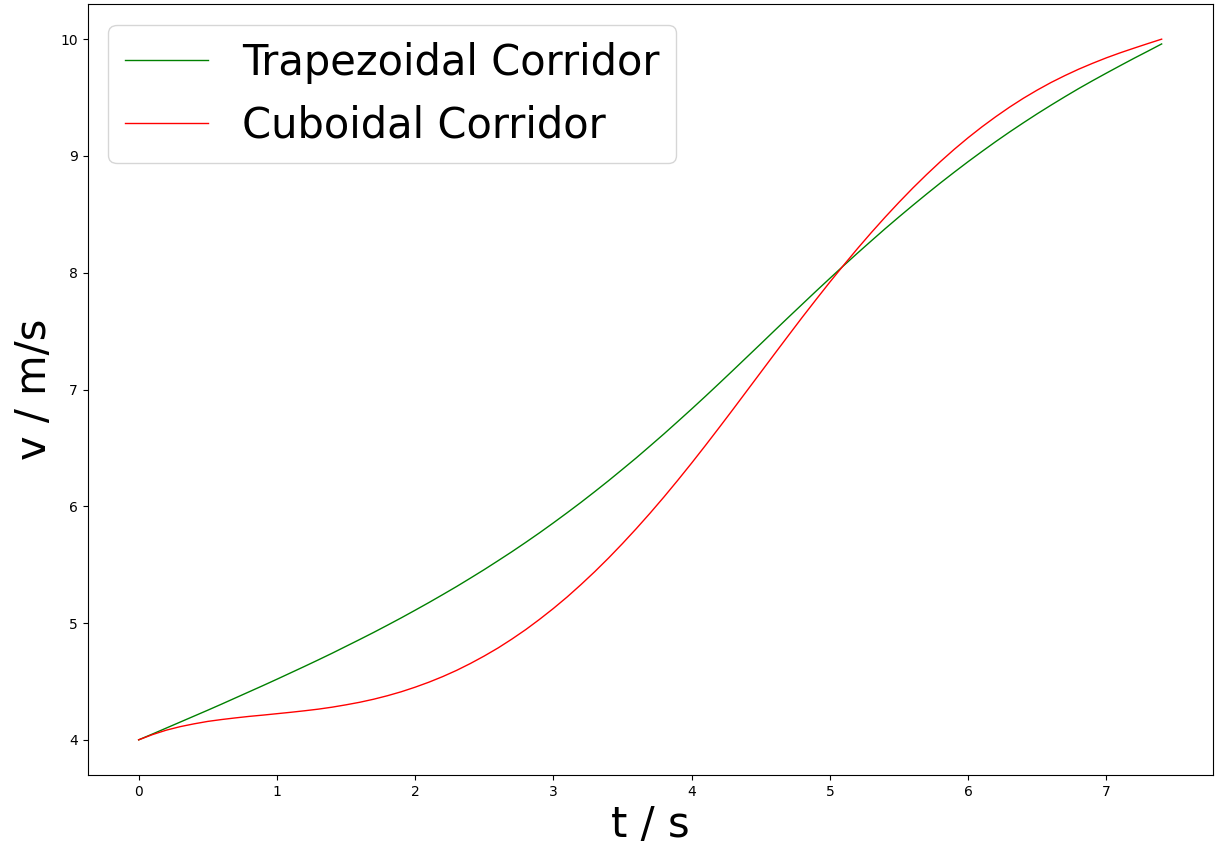}\label{fig:f10}}
    \subfloat[Lateral Velocity profiles]{\includegraphics[width=0.2\textwidth]{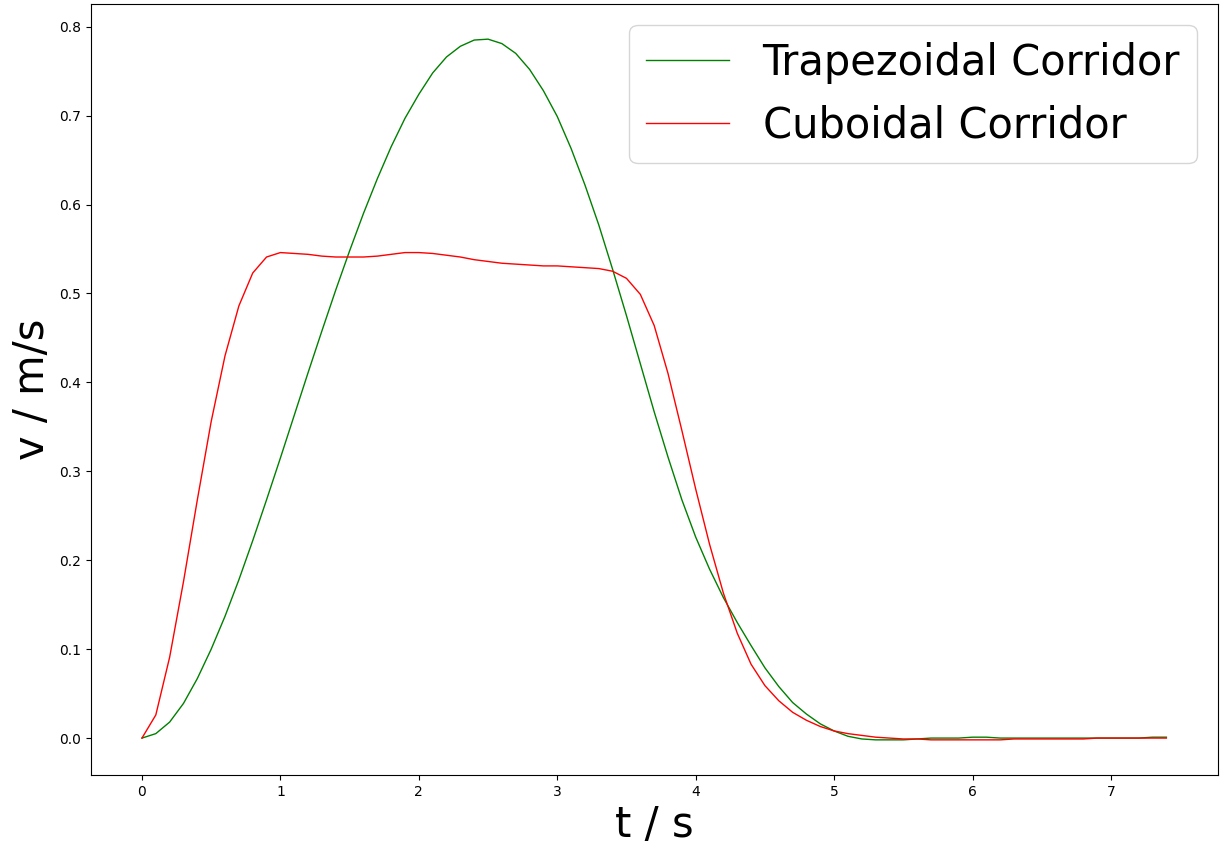}\label{fig:f11}}
    \caption{Piecewise Bézier polynomial and its dynamic profile}
    \label{fig: Fig 3}
\end{figure}

Fig. \ref{fig:f8} and Fig. \ref{fig:f9} show B\'{e}zier curves generated by cuboidal (red,  \cite{ding_safe_2019}) and trapezoidal-prism (green, ours) corridors for the scenario presented in Fig. \ref{fig:motivation}. From Fig. \ref{fig:f12}, we observe that the maximum acceleration required for our method is less than that needed by the cuboidal corridors approach. The superiority of using trapezoidal corridors is more clear from Fig. \ref{fig:f13}, which records the lateral acceleration of both the methods. We observe that our approach yields a smoother acceleration plot with minimal jerk and the lower maximum acceleration. We also test the maximum initial conditions of both the methods for the same scenario to show the effect of the enlarged search space. While using trapezoidal corridors, we can generate a trajectory for $a_s = 2$ \text{m/s}\textsuperscript{2}, $v_s = 10.5$ \text{m/s}, $a_l = 1.2$ \text{m/s}\textsuperscript{2}, $v_l = 2$ \text{m/s} where the bounds on longitudinal acceleration are $[-3, 2]$ \text{m/s}\textsuperscript{2} and those on lateral acceleration are $[-2, 2]$\ \text{m/s}\textsuperscript{2}. Using the cuboidal corridors fails to generate a trajectory for these initial conditions and is only successful when the initial velocity in the longitudinal direction is reduced to 9\ m/s. 

\subsubsection{Overtaking a low-speed vehicle in front} 
\begin{figure}[htbp]
    \centering
    \subfloat[Longitudinal acceleration]{\includegraphics[width=0.22\textwidth]{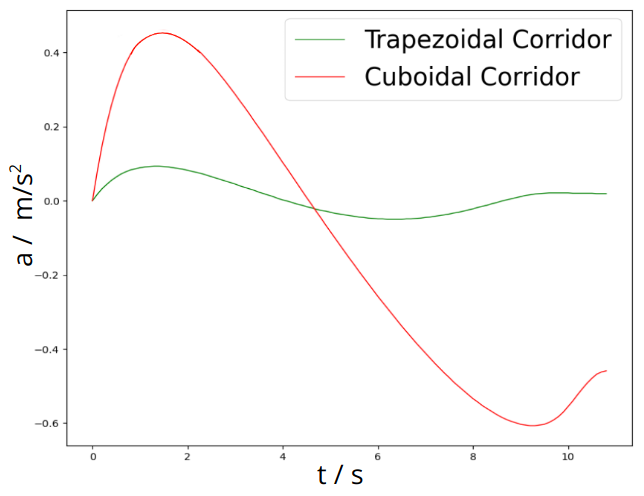}\label{fig:overtake_a}}\quad
    \subfloat[\textit{matplotlib} animation]{\includegraphics[width=0.23\textwidth]{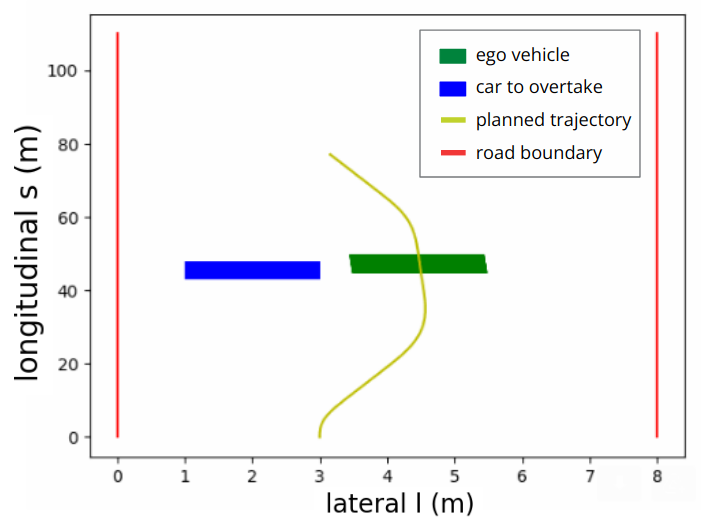}\label{fig:overtake_b}}
    \caption{Overtaking scenario}
    \label{fig:overtake}
\end{figure}

We test our planner on overtaking a slowly moving car in front by lane changing twice (second time to merge back into the original lane of the ego vehicle). In layered planning techniques, these kinds of scenarios are typically tackled by considering the obstacle to be static for a few seconds. Hence, this approach proves to be conservative. 
The differences in the longitudinal acceleration graphs between the two corridor generation techniques can be seen in Fig. \ref{fig:overtake}. Clearly, using the trapezoidal corridors generates a trajectory with much lower acceleration. Here, the vehicle in front is assumed to be moving with $v_s = 5$\ m/s and the ego vehicle's initial condition is $v_s = 7$ \ m/s.

\subsubsection{Unprotected left turn}
\begin{figure}[htbp]
    \centering
    \subfloat[ego vehicle is shown in blue with the corresponding reference trajectory]{\includegraphics[width=0.2\textwidth, angle=90]{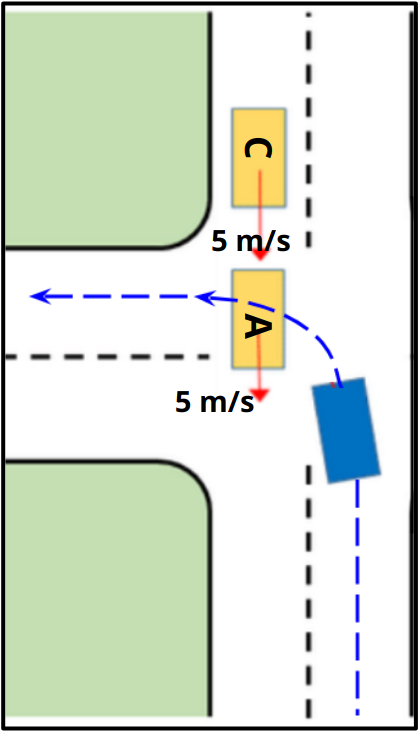}\label{fig:f18}}\\
    \subfloat[]{\includegraphics[width=0.2\textwidth]{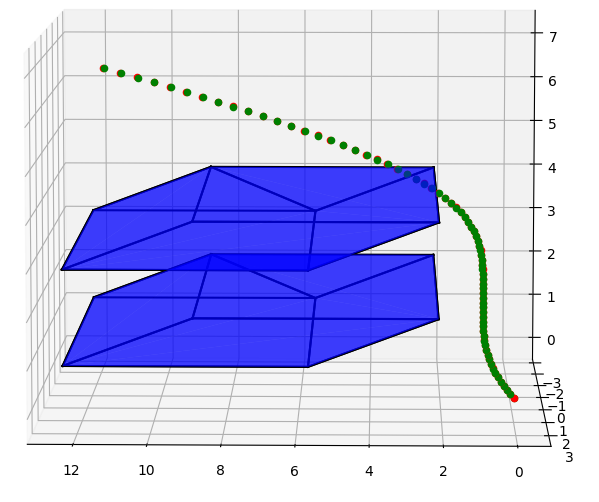}\label{fig:f16}}
    \subfloat[]{\includegraphics[width=0.2\textwidth]{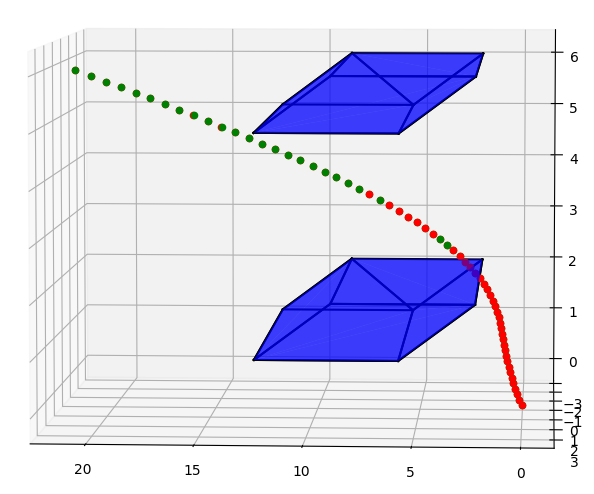}\label{fig:f17}}
    \caption{Unprotected Left Turn at an Intersection- trajectories obtained from both corridor construction methods (green: trapezoidal, red: cuboidal) are identical}
    \label{fig: Fig 5}
\end{figure}

As shown in Fig. \ref{fig:f18}, there are two cars coming from the front which obstruct the ego vehicle from making a left turn without yielding to them. As seen in Fig. \ref{fig:f16}, our planner can successfully find a trajectory while meeting all the safety and dynamic feasibility constraints. Since the ego vehicle needs to yield to the cars in front, we also test the maximum initial velocity ($v_s = 1$\ m/s) and acceleration ($a_s = 0.5$\ m/s\textsuperscript{2}) in the longitudinal direction for this case. If the distance between Car A and Car B is sufficient for the ego vehicle to go in between them, our planner finds the corresponding trajectory (Fig. \ref{fig:f17}). In this scenario, the additional search space obtained by trapezoidal corridors is not used at all, as the trajectory passing through the enlarged search space can only result in a lane change, which is not desired. Hence, both the trajectories obtained are the same and overlap each other, as seen in Fig. \ref{fig: Fig 5}.

\subsection{CommonRoad Simulations}
\begin{figure}[htbp]
    \centering
    \subfloat[$t=0\ s$]{\includegraphics[width=0.4\textwidth, angle=90]{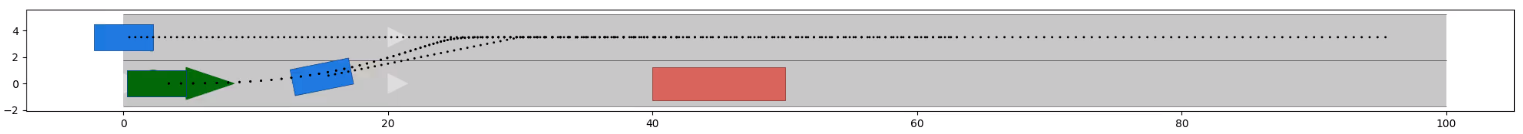}\label{fig:c1}}\quad\quad
    \subfloat[$t=2\ s$]{\includegraphics[width=0.4\textwidth, angle=90]{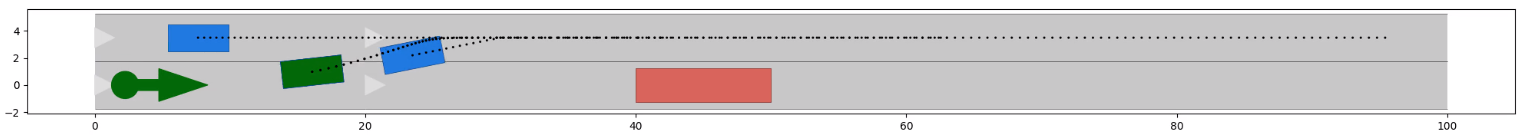}\label{fig:c2}}\quad\quad
    \subfloat[$t=4\ s$]{\includegraphics[width=0.4\textwidth, angle=90]{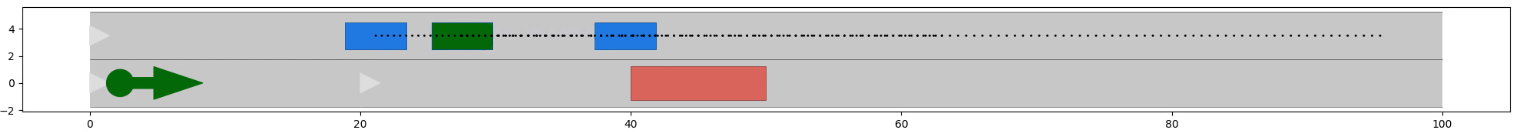}\label{fig:c3}}\quad\quad
    \subfloat[$t=7\ s$]{\includegraphics[width=0.4\textwidth, angle=90]{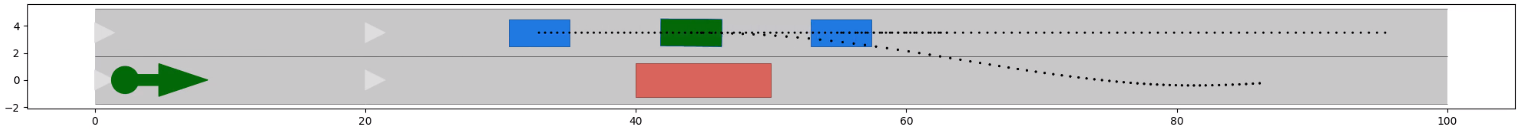}\label{fig:c4}}\quad\quad
    \subfloat[$t=9\ s$]{\includegraphics[width=0.4\textwidth, angle=90]{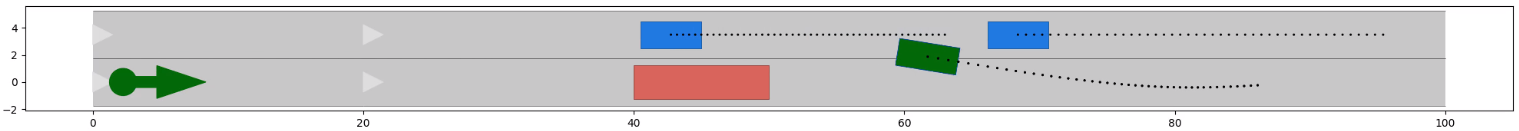}\label{fig:c5}}\quad\quad
    \subfloat[$t=12\ s$]{\includegraphics[width=0.4\textwidth, angle=90]{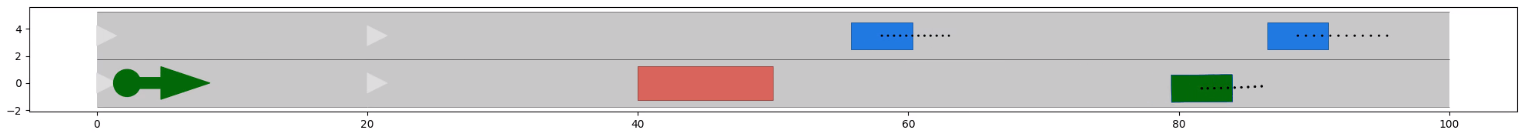}\label{fig:c6}}
    \caption{CommonRoad Simulation for ``Merging into another lane due to road construction" scenario. The ego vehicle (initial location is represented by the green arrow) is shown in green, and the other cars are shown in blue. The construction site (static obstacle) is shown in red.}
    \label{fig: Fig 6}
\end{figure}

The simulations in this part are conducted on the CommonRoad platform \cite{commonroad}, which provides an interactive simulated and non-interactive real traffic environment for validating motion planning algorithms. A given scenario is considered ``solved” when the ego vehicle reaches the desired goal region while satisfying all the constraints. We visualize the bird's-eye view simulation of the lane change scenario in Fig. \ref{fig:motivation}. The results obtained using our approach can be seen in Fig. \ref{fig: Fig 6}. The cuboidal corridor approach did not yield a collision-free trajectory as it failed to replan owing to the lack of search space. It also had significantly high acceleration as can be observed in Fig. \ref{fig:f12}.

\section{CONCLUSION}

In this paper, we propose a novel convexification algorithm for generating safety corridors in the $S-L-T$ space. We show that our method of trapezoidal prism-shaped corridors enlarges the solution space as compared to the existing cuboidal corridors-based method. We provide the sufficient conditions of control points in the trapezoidal corridors to provably guarantee the safety of trajectories represented by B\'{e}zier polynomials. Finally, we formulate the trajectory optimization as a QP problem. The numerical and CommonRoad simulations show that the proposed approach is superior in terms of optimality and low failure rates. Future work includes using a dynamic programming-based approach to generate a comfort-optimal reference trajectory in the $S-L-T$ space.


\bibliographystyle{IEEEtran}
\bibliography{./IEEEfull,refs}

\end{document}